\documentclass[sigconf,screen]{acmart}
\usepackage{multirow}
\usepackage{subcaption}
\usepackage{color}

\renewcommand\footnotetextcopyrightpermission[1]{}

\AtBeginDocument{%
  \providecommand\BibTeX{{%
    \normalfont B\kern-0.5em{\scshape i\kern-0.25em b}\kern-0.8em\TeX}}}

\acmSubmissionID{1884}

\begin{document}

\title{Shifting Perspective to See Difference: A Novel Multi-View Method for Skeleton based Action Recognition}


\author{Ruijie Hou}
\authornote{Both authors contributed equally to this research.}
\affiliation{%
  \institution{Zhejiang University}
  \city{Hangzhou}
  \country{China}
  }
\email{ruijie.hou@zju.edu.cn}

\author{Yanran Li}
\authornotemark[1]
\affiliation{%
  \institution{Zhejiang University}
  \city{Hangzhou}
  \country{China}
  }
\email{buliyanran@gmail.com}
  
\author{Ningyu Zhang}
\affiliation{%
  \institution{Zhejiang University}
  \city{Hangzhou}
  \country{China}
  }
\email{zhangningyu@zju.edu.cn}

\author{Yulin Zhou}
\affiliation{%
  \institution{Zhejiang University}
  \city{Hangzhou}
  \country{China}
  }
\email{zhou.yulin@zju.edu.cn}

\author{Xiaosong Yang}
\affiliation{%
  \institution{Bournemouth University}
  \city{Poole}
  \country{United Kingdom}
  }
\email{xyang@bournemouth.ac.uk}
  
\author{Zhao Wang}
\authornote{Corresponding Author}
\affiliation{%
  \institution{Zhejiang University}
  \city{Hangzhou}
  \country{China}
}
\email{zhao\_wang@zju.edu.cn}
% \renewcommand{\shortauthors}{Ruijie Hou, et al.} 
\renewcommand{\shortauthors}{Ruijie Hou et al.}

%  \renewcommand{\shorttitle}{Shifting Perspective to See Difference} 

\begin{abstract}
  Skeleton-based human action recognition is a longstanding challenge due to its complex dynamics. Some fine-grain details of the dynamics play a vital role in classification. The existing work largely focuses on designing incremental neural networks with more complicated adjacent matrices to capture the details of joints relationships. However, they still have difficulties distinguishing actions that have broadly similar motion patterns but belong to different categories. Interestingly, we found that the subtle differences in motion patterns can be significantly amplified and become easy for audience to distinct through specified view directions, where this property haven't been fully explored before. Drastically different from previous work, we boost the performance by proposing a conceptually simple yet effective Multi-view strategy that recognizes actions from a collection of dynamic view features. Specifically, we design a novel Skeleton-Anchor Proposal (SAP) module which contains a Multi-head structure to learn a set of views. For feature learning of different views, we introduce a novel Angle Representation to transform the actions under different views and feed the transformations into the baseline model. Our module can work seamlessly with the existing action classification model. Incorporated with baseline models, our SAP module exhibits clear performance gains on many challenging benchmarks. Moreover, comprehensive experiments show that our model consistently beats down the state-of-the-art and remains effective and robust especially when dealing with corrupted data. Related code will be available on \url{https://github.com/ideal-idea/SAP}.

\end{abstract}



\begin{teaserfigure}
\centering
  \includegraphics[width=\textwidth]{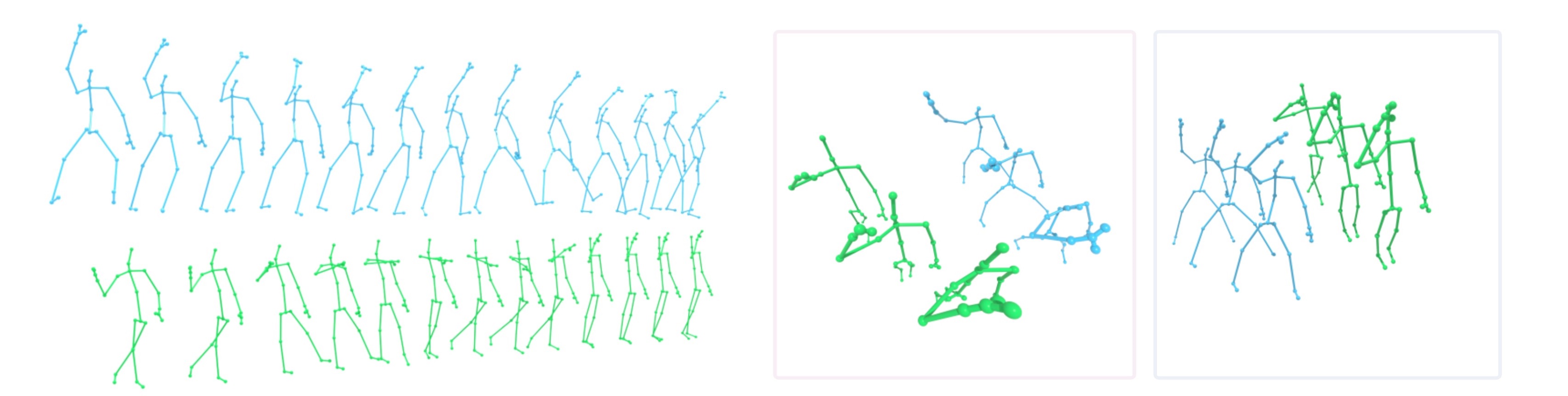}
  \caption{The action sequences (left, blue stands for \textit{waving hands}, green stands for \textit{drinking water}) have similar appearance but different meaning, could be distinguished easily from specified view (middle and right)}
  \Description{We depict the different observations under different views for the same action sequences. And this figure illustrates the discriminative features under some views for similar action sequences.}
  \label{fig:teaser}
\end{teaserfigure}

\maketitle

\section{Introduction}
\label{sec:intro}
Human action recognition is a crucial topic in multi-media processing due to its significant role in real-life applications~\cite{poppe2010survey}, such as video surveillance~\cite{khan2020human}, human-robotics interaction~\cite{lee2020real}, health care~\cite{lopez2019human} and autonomous driving~\cite{camara2020pedestrian}. The mainstream work recognizes actions from three types of input data -- RGB videos, depth RGB videos or 3D skeletons. With the development of low-cost motion sensors and effective human pose estimation techniques, skeleton-based action recognition becomes a highly active area recently because they are more efficient, compact and robust in human-centred scene understanding compared with its video counterpart. Under complex appearance variations such as background distractions and illumination changes, skeleton-based representation is not affected and conveys relatively high-level information.

The key challenge of skeleton-based actions recognition is to learn discriminated features for classification. For this purpose, a variant Recurrent Neural Network (RNN) based~\cite{du2015hierarchical,zhang2019view} or Convolutional Neural Network (CNN) based models~\cite{du2015skeleton,li2017skeleton} are proposed in the early stage. They either regard the motion data as vector sequences or pseudo-images, which exploits the joint relationships inadequately. In contrast to them, graph convolutional network (GCN) based models~\cite{yan2018spatial, li2019spatio, si2019attention, shi2019skeleton, xia2021multi, chen2021multi} formulate spatial-temporal graphs to model human actions, which naturally fit the data structure and achieve superior performance. For example, ST-GCN~\cite{yan2018spatial} constructed a spatial-temporal skeleton graph which takes the joints as graph nodes, natural connections of body and time as edges. Thus, an adjacency matrix of the skeleton graph is built that contains spatial-temporal relations between joints. Following their work, the mainstream approaches~\cite{li2019spatio, si2019attention, shi2019skeleton} are focusing on tailoring the design of adjacency matrix to capture more fine-grain details. For instance, some of their work consider the different scale of receptive field~\cite{cheng2020skeleton}, capturing both the short-term trajectory and long-term trajectory
~\cite{chen2021multi}, or design multi-scale ST-GCN~\cite{xia2021multi}. Another group of trending approaches~\cite{shi2019two, song2022constructing, yu2022multi} leverage the capacity of GCN models by introducing multiple streams. They usually take joints (relative and absolute), velocity (joints motion and bones motion) and bones as inputs and fuse all the features together as the final representation.

Despite their tremendous effort, the state-of-the-art models are still prone to produce wrong classifications in some broadly similar actions. For instance, the drinking water action and waving hand action contains very similar motion patterns on their legs and relations between arms. However, human beings can still distinguish them due to the small local details around the arm area. Although the existing work elaborates sophisticated incremental modules to capture these details, the learned features of the two actions are too ambiguous and confused for the classifier since they contain too much similarity and the distinctive fine details only take a low percentage of the information. Learning a proper decision boundary is difficult in these scenarios, since the action data points may have a high inter-class similarity. In contrast, we notice that human users can easily distinguish the two actions by rotating the sequences or changing the view angles. In this way, the discriminated characteristics of actions can be more dominated and distinctive for recognition when the 3D skeletons are viewed from a certain direction. As shown in Figure~\ref{fig:teaser}, the action sequences can be viewed from different angles to present more distinctive motion patterns which make them much easier to be classified. Therefore, we claim that the translations of actions through different views will offer more effective and distinct dynamic patterns so that the model is more capable to deal with the intractable action classification problem.

In the literature, the multi-view strategy has demonstrated an effective enhancement in the 3D shape recognition task~\cite{hamdi2021mvtn, su2015multi}. However, few approaches for skeleton-based action recognition have deeply investigated the problem from this perspective. Zhang \textit{et al.}~\cite{zhang2019view} attempt to address the view variation problem by introducing a CNN-based view adaption network. Their work validates that transforming the data through views will bring performance gains. However, their work mainly focus on avoiding the motion direction variation problem rather than exploiting the information from different views. To date, the problem of how to utilise view features to leverage the performance of action recognition hasn't been thoroughly pursued and carefully investigated to our best knowledge. 

Motivated by this insight, we delve deeper into the idea of enhancing feature representation through exploiting multiple views and elaborate the special view streams beside the ordinary joints, bones and velocity streams for the action recognition framework. To achieve this goal, we need to solve two imperative problems: how to represent the inputs under each view and how to learn the proper view angles. For the first issue, we design a novel Angle Representation to translate the coordinate-based 3D skeleton into a triplet anchor-defined structure for each view. More specifically, our view is determined by two anchor points in the 3D Euclidean space. The triplet is consist of the two anchor points and any joint from the human skeleton. A unique angle value could be obtained from each triplet so that the human pose can be defined by a set of angles with determined pair of anchor points. Interestingly, the current streams are either defined by single joints or binary pairs of points, but our view stream is defined by triplets of points. 
To address the second question, we delicately design a novel skeleton-anchor proposal (SAP) module to learn the position of anchor points. Since attention has been widely used as a powerful mechanism to exploit the relationship between the joints, we design the SAP module as a variant multi-head attention which takes each pose as input and outputs multiple anchor pairs. For further analysis, our SAP is designed to contain an upgrade ability that can control the anchor locations around the body or inside the body. 
Towards this end, our multi-view stream for skeleton-based action recognition is built up. We will put the initial action data into our SAP to learn multiple anchor points. Then the angle represented action can be determined. After that, each angle represented action is fused and sent into the baseline module to extract the feature representations. 

Notably, the baseline module can be implemented by any state-of-the-art skeleton-based action classification model. Here we validate our idea by implementing the GCN-based model MS-G3D~\cite{liu2020disentangling} and CNN-based model VA-CNN~\cite{zhang2019view}. To verify the superiority of the proposal idea, extensive experiments are conducted on the challenging benchmark dataset NTU-RGB+D~\cite{shahroudy2016ntu}. Our model significantly outperforms state-of-the-art works in extensive evaluations. Moreover, our model demonstrates remarkable improvements even on corrupted and noisy 3D skeleton action data, which reveals that the multi-view strategy can enhance the robustness of action recognition. 
The contributions of this work are summarized as follows:
\begin{itemize}
  \item For the first time, we investigate the multi-view strategy in-depth and propose a novel view stream for the skeleton-based action recognition problem.
  \item We design a novel multi-head skeleton-anchor proposal module (SAP) and angle representation. Our experiment demonstrates that view features offer more distinctive and complementary information for recognition.
  \item We conduct comprehensive evaluations on the selection of the views and fusion strategies of multi-view features. Fruitful insights are provided for the field. Notably, our method is generic and robust to deal with noisy data, which is a very common issue for skeleton-based action recognition.  
\end{itemize}

\begin{figure*}
  \includegraphics[width=0.8\textwidth]{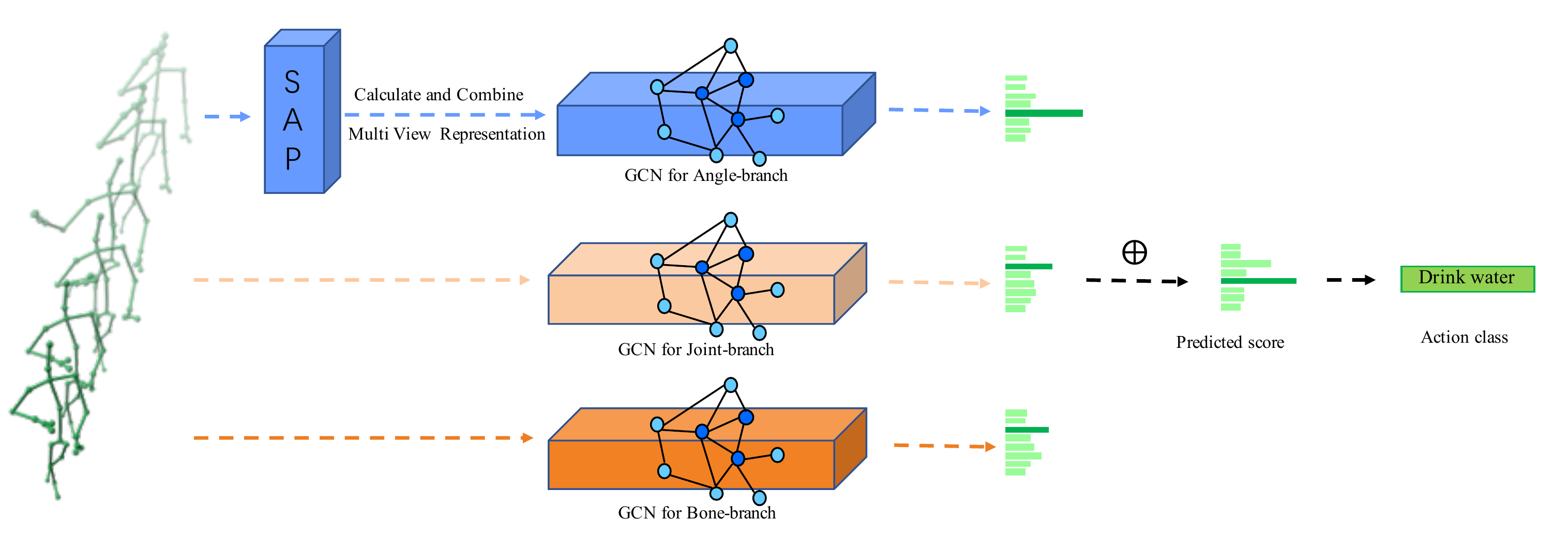}
  \caption{The detailed schematic diagram of our Multi-view framework for skeleton-based action recognition}
  \Description{We illustrated the pipeline of our multi-view. We adopted the ensemble strategy to get the strength of the view stream.}
  \label{fig:pipeline}
\end{figure*}

\section{Related Work}
\label{sec:rw}
\subsection{Skeleton-based action recognition}
Action recognition based on skeleton data has received lots of attention due to its robustness and efficiency. Handcrafted features were used in early approaches, where features could be manually designed based on joint angles~\cite{ofli2014sequence}, kinematic features~\cite{zanfir2013moving}, trajectories ~\cite{wang2016graph} or their combinations~\cite{wang2016adaptive}. With the development of deep learning, many CNN or RNN based data-driven methods that could automatically learn the action patterns have been proposed. For instance, the skeleton action could either be treated as motion images in CNN approaches~\cite{ke2017new,kim2017interpretable} or modelled as sequences of coordinates in RNN approaches~\cite{du2015hierarchical,liu2016spatio}. For instance, a hierarchical bidirectional RNN has been employed to capture dependencies within body parts~\cite{du2015hierarchical}. A trimmed skeleton sequence has been used in a CNN architecture for action classification~\cite{du2015skeleton,li2017skeleton}. However, the aforementioned methods fail to fully 
exploit the inherent relationships between joints since the connectivity of the human body skeleton is very different from languages and images.

\begin{figure}[b]
  \centering
  \includegraphics[width=0.45\textwidth]{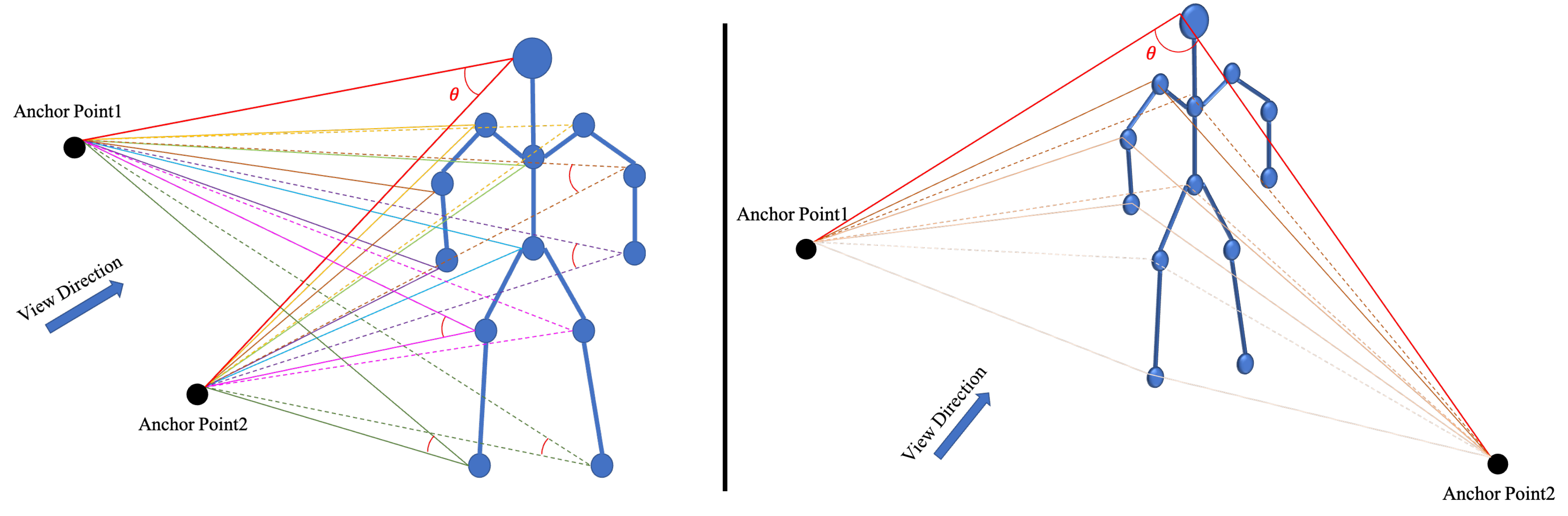}
  \Description{We illustrate under the different views determined by the different anchor pairs, we get the angle representation by calculating the cosine of the angle of the two edges from the joints to the anchors.}
  \caption{The angles under different views}
  \label{fig:Angle}
\end{figure}

Graph-based methods have sparked a revolution in skeleton-based action recognition studies recently. The first GCN model for skeleton-based action recognition is ST-GCN proposed by~\cite{yan2018spatial}, where the skeleton is treated as a graph, with joints as nodes and bones as edges. Following the work of ST-GCN, a number of approaches explored the relationship between distant joints~\cite{shi2019skeleton,li2019spatio,zhang2020context} to increase the information. In addition, multi-scale structural feature representation methods have been developed via higher-order polynomials of the skeleton adjacency matrix. For instance, a multiple-hop module is used to break the limitation of representational capacity caused by first-order approximation~\cite{peng2020learning,liu2020disentangling,xia2021multi}. Inspired by~\cite{liu2020disentangling}, a sub-graph convolution cascaded by residual connection with enrich temporal receptive field is introduced by~\cite{chen2021multi}. A combination approach called Efficient GCN is designed by~\cite{song2022constructing}. A multi-granular GCN based method on the temporal domain is designed by~\cite{chen2021learning}. Angle information extracted from manually specified joint groups is fused to GCN model in~\cite{qin2021fusing}. These researches typically introduce incremental modules to increase the information of finer details. However, few of them pay attention to research on the multi-view feature representation of the skeleton-based action data. Normally, most of the existing work proposes multi-stream pipeline to strengthen the model's ability to learn more expressive features. For example,  2s-AGCN~\cite{shi2019two} utilised joints and bones input in their two-stream framework. EfficientGCN~\cite{song2022constructing} considered joints, bone and velocity to increase the capacity of their model. Furthermore, joints motion and bones motion are added by ~\cite{yu2022multi} as extra streams in the feature learning pipeline. However, the view stream has not been formally proposed and is well designed in the field due to our best knowledge. In this work, we fill this gap by introducing a multi-view stream that learns adaptive view angles to increase the model competence.

\subsection{Multi-view strategy for 3D objects}
% Third Paragraph about Multi-view Projection in 3D vision
Utilizing a series of 2D View images to categorize 3D objects has been widely studied in recent years. The first milestone approach is MVCNN~\cite{su2015multi}, which extracts the CNN features for each view to classifying the 3D objects. They reveal that even a single view feature can beat down the other 3D descriptors. 
Following this work, many approaches~\cite{qi2016volumetric, johns2016pairwise, esteves2019equivariant, wei2020view, chen2018veram, hamdi2021mvtn} attempt to explore more effective view features and view selections. VERAM~\cite{chen2018veram} proposes an RNN-based Attention module to select the best views which are more informative and distant from each other. ViewGCN~\cite{wei2020view} investigates how to aggregate all the view descriptors into a global representation by constructing a view graph. In contrast to these fixed viewpoints strategies, MVTN~\cite{hamdi2021mvtn} introduces a transformation-based network to learn adaptive viewpoints for any specific 3D vision tasks. Inspired by their great success, we propose the first multi-view base skeleton based action recognition work in this paper. Different from any 3D approaches, we design a novel view anchors learning and selection strategy specifically for skeleton data.

\subsection{View learning for skeleton-based actions}
Due to date, the multi-view strategy has never been well noticed for skeleton-based action recognition. The existing work only treats view transformation as a pre-procession strategy to ensure the view-invariant property in the classification task. 
In the early stage, most of them~\cite{du2015hierarchical, jiang2015informative, li2017adaptive, liu2016spatio, liu2017global, shahroudy2016ntu, wang2013learning} align the body orientation of the whole sequence to a certain direction, which may produce weird actions. Further, Zhang \textit{et al.}~\cite{zhang2019view} proposed the first work which learns adaptive viewpoints based on different motion content. They design RNN and CNN based modules to determine the observation direction for each sequence and translate the skeleton data according to new viewpoints. Their experiments validate that the view translation is effective to leverage the recognition accuracy. However, they only limited the strength of views for orientation normalization and their views learning methods are restricted for CNN and RNN architectures. Moreover, only one viewpoints are determined for each time slot in their model. To the best of our knowledge, none of the aforementioned works discusses how to select multiple adaptive views for the recognition task. In contrast, we delve much deeper into the multi-view strategy and carry out a comprehensive study for multiple view angles selection in our work. 
%-------------------------------------------------------------------------

\section{Methodology}
\label{sec:meth}
% the motivation & summarize of what we do in this paper
Under different camera view directions, the 3D skeleton represented action sequences show very different visualization and characteristics. As illustrated in Figure \ref{fig:teaser}, action sequences of drinking water could be similar to sequences of waving hands via front view. On the contrary, the discriminated part of the posture of drinking water is able to be obtained from other views, which could highlight the difference between the moving hands significantly. Inspired by that, we introduce a new multi-view strategy that recognizes the action from a series of different views, where both global and local discriminate information can be explored. In this section, we present the overview of the proposed multi-view framework, the representation of view and the view estimation module.

\subsection{Overview of the Multi-view framework}
% Overview of the pipeline
The detailed schematic diagram of our multi-view framework for skeleton-based action recognition is presented in Figure \ref{fig:pipeline}. Firstly, the input action sequence $S$ will be processed into three streams -- angles, joints and bones. The joints and bones streams are typically fed in the GCN-based models following a similar design of MS-G3D \cite{liu2020disentangling}. The angle stream is designed to introduce multiple representations of the action under multiple views. More concretely, we firstly learn $M$ different views for each sequence. Then a set of action representations $\{S_i,|\ i = 1, ..., M\}$ can be obtained by transforming the original action sequence into an angle representation. All of these action sequences $\{ S, S_i| \ i = 1, ..., M \}$ are combined and fed into the following GCN-based models. Here we adopt the state-of-the-art graph neural networks for action recognition as the feature learning backbones. After that, all the features of different streams are aggregated together and sent into the last softmax layer for the action classification task. Notably, the backbone network can be easily replaced with another action classification model. The angle representation and the view estimation module are comprehensively 
introduced in the following paragraph.

\subsection{Angle Representation for each action}
One of the most important questions to use the multi-view strategy for skeleton data is how to represent the transformation of actions under a certain view. The existing multi-view approaches in 3D vision utilize the projection image of the 3D objects under certain views and extract features through various CNN modules. For skeleton-based action recognition, the existing work~\cite{zhang2019view} trains the network to learn a set of Euler angle $\alpha, \beta, \gamma$ and use the rotation matrix to translate the orientations of the human pose. Such a method is convenient for view normalization but not convenient for multi-view feature learning. Moreover, calculating the projections of skeletons for different views is tedious and unnecessary. Hence, we propose a straightforward angle representation to translate the action skeleton under a certain view uniquely with a pair of learned anchors. Compared with the projection method, our angle representation is more intuitive and expressive. The mathematical formulation of our angle representation in a certain view is described in the following.

We denote the input action sequence as $S$ and $T$ frames are contained in $S$. Each frame is formulated as a human graph $S^t$ which is consisted of joints and bones. Each frame can be written as $\{(V^t_i, E^t_j) \}$. The $i$th joint in $t$th frame is a coordinate $(x^t_i, y^t_i, z^t_i)$ in the 3D space and denoted as $V^t_i$. The $E^t_j$ denotes the connectivity of the $jth$ limb. In our work, we assume the location of view is 360 degrees free to capture the most useful information. As shown in Figure \ref{fig:Angle}, every view is defined by a pair of anchor points. Once a pair of anchor points are determined, the angle representation in this view $S_{\Theta}$ of action $S$ is produced. 

Each blue node is a joint on the human body skeleton at frame $t$. The coordinate of the joint is denoted as $C^t = <x, y, z>$. The two black nodes are the anchor points and can be written as $A = <a_x, a_y, a_z>$ and $B = <b_x, b_y, b_z>$. The cosine of an angle $\theta$ for the joint $C^t$ could be obtained through Equation \eqref{equ:gen_angle}. 
To this end, the problem of certain view estimation is transferred to the task of determining its corresponding pair of view anchors.  

\begin{equation}
\label{equ:gen_angle}
    cos(\theta^t)=\left\{
\begin{array}{ll}
\frac{\overrightarrow{AC}^t \cdot \overrightarrow{BC}^t}{\|\overrightarrow{AC}^t\|\|\overrightarrow{BC}^t\|} & \text { if } A \neq C \text{ and } B \neq C, \\

0  & \text {if } A = C \text { or } B = C .
\end{array}\right.
\end{equation}

Therefore, all the joints on the human body skeleton can be mapping to a certain cosine of the angle $\theta$. Summarily, the following view translation through angle representation can be formulated:
\begin{equation}
    \text{View Translation:  } S^t=\{(V^t_i, E^t_j) \} \longmapsto S_{\theta}^{t}=\{(\Theta^t_i, E^t_j) \}
\end{equation}
More concretely, $S$ is the joint stream and $S_\theta$ is the angle stream.

Each pair of anchor points will generate a new angle sequence $S_{\theta}$. All the angle sequences will be combined and fed into the GCN-based action recognition models for feature learning similar to $S$. In this way, we obtain the feature representation for each skeleton-based action sequence data from different views.

Generally, we learn the set of different views by determining pairs of anchor points and transferring the original coordinate represented skeleton into a series of angle representations. The details of anchor points generation are described in the next section. 

\begin{figure*}
  \includegraphics[width=0.9\textwidth]{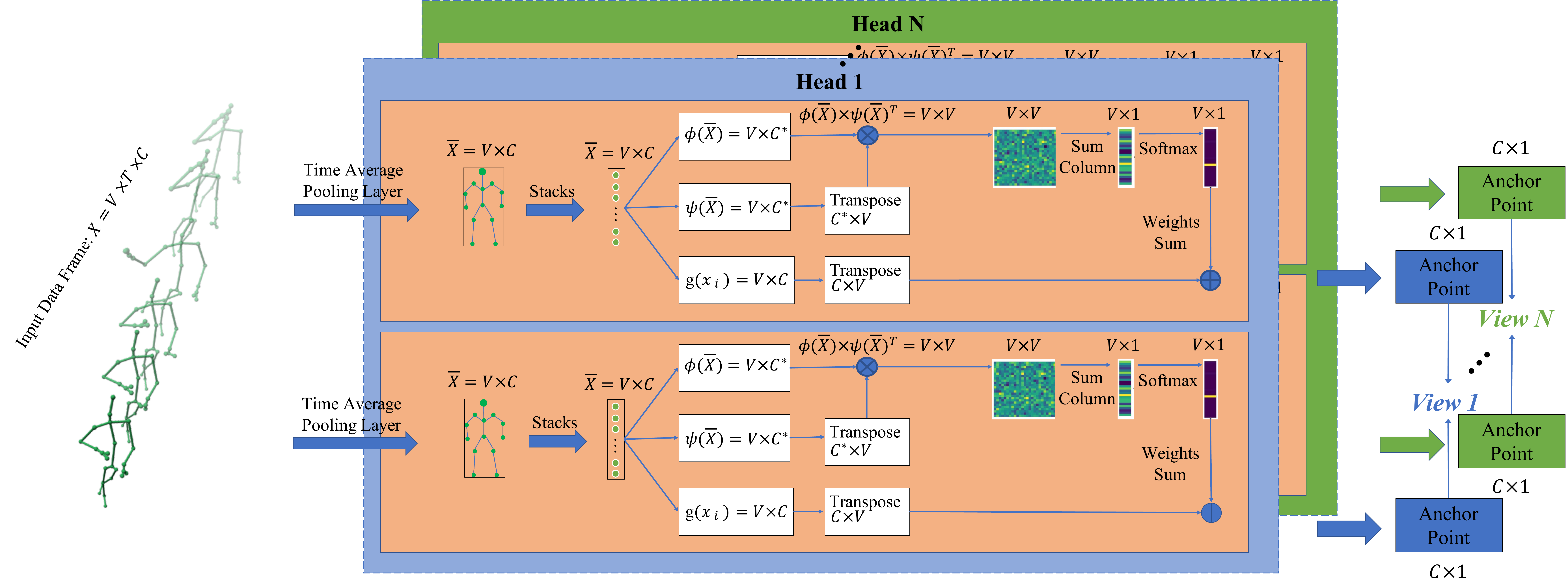}
  \caption{The diagram of our proposed Skeleton Angle Proposal (SAP) module for view anchors learning. We first perform time average pool on input data. Then pass the data to three linear transformation to get the Q,K,V respectively. After getting the similar matrix, we get the sum along the row axis and perform the softmax function. Then, we get the anchor point location through the weighed sum. Then we get multi views determined by multi pair of anchors}
  \Description{We illustrated the detail of the SAP.We first perform time average pool on input data. Then pass the data to three linear transformation to get the Q,K,V respectively. After getting the similar matrix, we get the sum along the row axis and perform the softmax function. Then, we get the anchor point location through the weighed sum. Then we get multi views determined by multi pair of anchors.}
  \label{fig:SAP}
\end{figure*}

\subsection{SAP module for multi-view anchor points learning}
% motivation to design SAP and limitation of the existing method
In order to determine the multiple pairs of anchor points for a given action sequence, we elaborated a Skeleton-Anchor Proposal (SAP) module. The previous work~\cite{zhang2019view} learn one viewpoint by CNN-based and RNN-based frameworks. They either use the hidden state vector at time $t$ as input or form the skeleton as a pseudo-image to learn the viewpoint parameters. However, such work has not paid attention to exploring joints relationship information. Moreover, another important limitation of such viewpoint learning methods, is that they are tackling the motion direction variation problem rather than exploiting the view information . In contrast, our module employs a modified self-attention mechanism which is more effective to exploit joint relationships and flexible to control and manipulate multiple pairs of anchor points to generate multiple views. The overview of the proposed Skeleton-Anchor Proposal (SAP) module is shown in Figure \ref{fig:SAP} and detail would be discussed in the following. 

The attention mechanism performs typically in three steps: (1) Calculate the alignment score for every element (feature vectors). (2) Compute the weights for every element from softmax. (3) Generate a unique vector $\vec{A}$ by summing up all the elements. 

In our design, we set the elements $\bar x_i$ to be the average value of the coordinates of the $i$-th joints in all frames. The following Equation~\eqref{equ:x} gives out the math formulations:

\begin{equation}\label{equ:x}
    \bar x_i = \frac{\sum_{t=1}^{t=T}x_{i,t}}{T}
\end{equation}
where $\bar x_i\in \mathbb{R}^3$ and $T$ is the number of frames for each action sequence. After that, we will calculate the alignment score and weights by the following Equation~\eqref{equ:score}.
\begin{equation}\label{equ:score}
\begin{aligned}
a_i = \sum_{\forall{j}}\phi(\bar x_i)\times\psi(\bar x_j)^T \\
weight_i = softmax(a_i) 
\end{aligned}
\end{equation}
In the equation, $\phi$ and $\psi$ are two networks and their parameters are trained during learning. Finally, the position obtained by the weighted sum of all elements is used as the anchor point, which is defined as Equation~\eqref{equ:center}:
\begin{equation}\label{equ:center}
    center = \sum_{\forall i}weight_i \times g(x_{i})
\end{equation}
The $g(x_{i})$ is used to determine the range of anchor locations, which would be discussed in the following paragraph. The module described above is used to generate an anchor point. We add multiple pairs of this module to generate multiple pairs of anchors similar to the multi-head structure in self-attention. After that, we use these pairs of anchors and action sequence $S$ to generate angle representation for different views with Equation~\eqref{equ:gen_angle}. We will further introduce the necessity of multi-view and the principle of choosing view anchors in the following sections.

\begin{figure}[b]
  \includegraphics[width=0.4\textwidth]{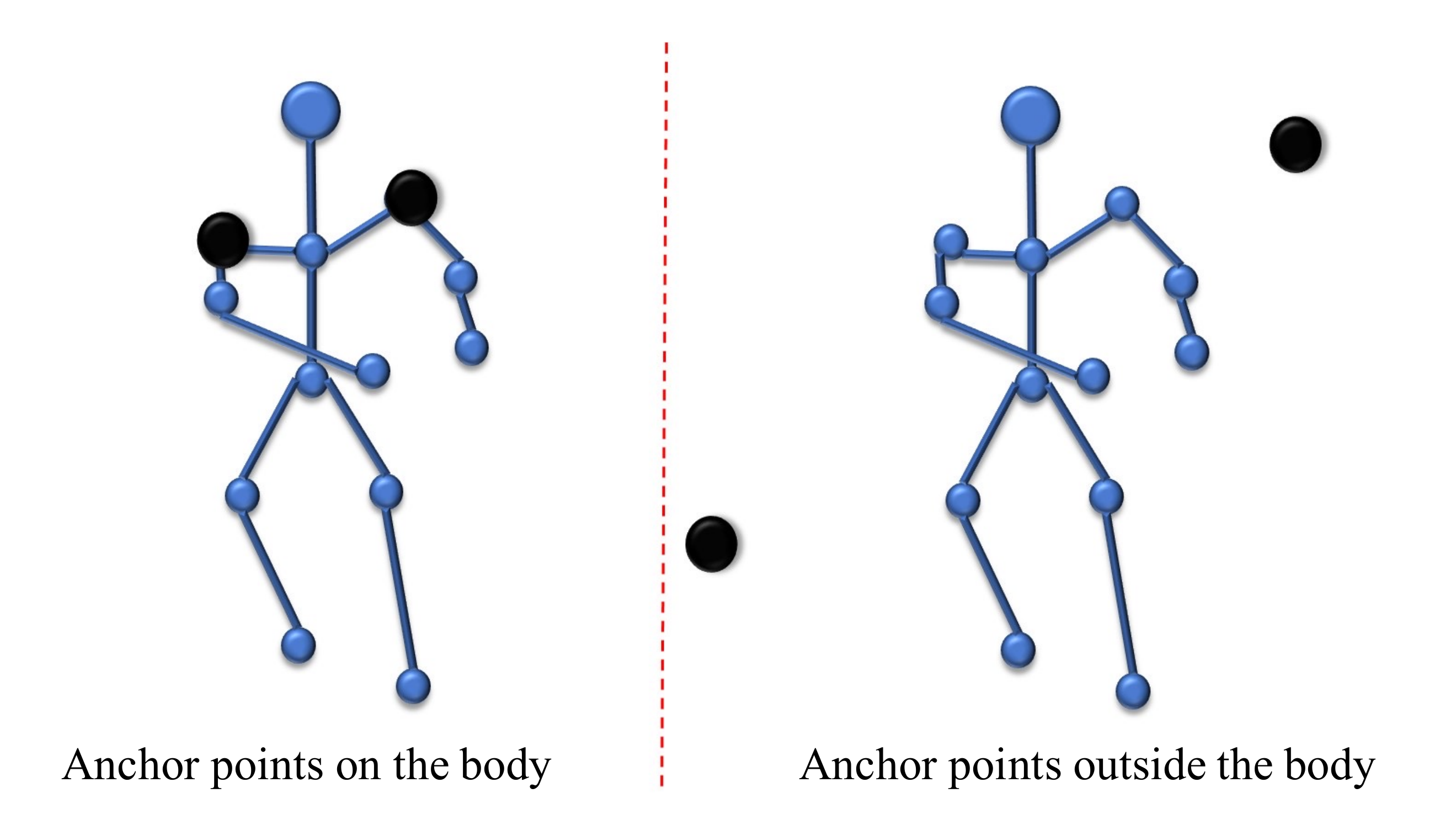}
  \caption{The anchor points selection }
  \Description{We illustrate the different anchor locations. There are two sub figures. The left figure's anchors are located in the joints. And the right figure's anchors are located in around the body.}
  \label{fig:AnchorPoints}
\end{figure}

\subsection{Multi-View anchor points selection}
Previous 3D object recognition works have shown that using a series of viewpoints to cover the entire object would provide a much better classification performance. We use a similar strategy that leverages discriminated information with multi-view since it's straightforward that some actions could be easily recognized under different views.  A multi-head structure in the SAP module is designed to learn multiple views for our task. Selecting a set of effective multiple views is not trivial and their range is essential. in addition, these selected views are supposed to maximise the scope of reception. This proposed SAP module provides a flexible way to adjust the position of anchor points in order to improve the scope of reception. 

As shown in Figure \ref{fig:AnchorPoints}, there are two choices for the view anchors' location: (1) the anchors are on the original body joints; (2) the anchors are located around the original body. This is achieved by the add a control component of $\alpha$ and $g$ in the anchor generation function. The formulation is given in the following:

\begin{equation}\label{equ:Control}
\begin{aligned}
a_i = \sum_{\forall{j}}\phi(\bar x_i)\times\psi(\bar x_j)^T \times \alpha
\end{aligned}
\end{equation}

In Equation~\eqref{equ:Control}, $\alpha$ is used to control the degree of dispersion of the softmax generation weight. When $\alpha$ increases, the maximum value of softmax will be increased and the minimum value will be decreased. In this way, the generated anchor point is encouraged to be located on an origin joint.

\paragraph{The anchor points location}

Here a linear function $g(x)=x$ is used in this work, When $\alpha$ goes large, the weighted sum can achieve the effect of selecting the most salient node among the original nodes, which can make the generated anchors fall on the original joints. We illustrated the selection of anchor points on the left of Figure~\ref{fig:AnchorPoints}. When the value of $\alpha$ goes small, the generated anchors will fall within the body.  In addition, the anchors generated by self-attention could locate out of the natural body range, which can be selected arbitrarily in the entire coordinate system. To achieve that, we set $g(x_{i})=w_g \times \bar x_i$, where $w_g$ stands for a linear fully connection that is similar to $\phi(x)$ in Equation~\eqref{equ:x}. Such linear transformation operation, implemented by a full connection with a 1x1 convolution channel, can make the anchors generated by SAP fall locate around the body.

\subsection{How to aggregate the views}
\begin{figure}[htbp]
    \centering
    \includegraphics[width=0.45\textwidth]{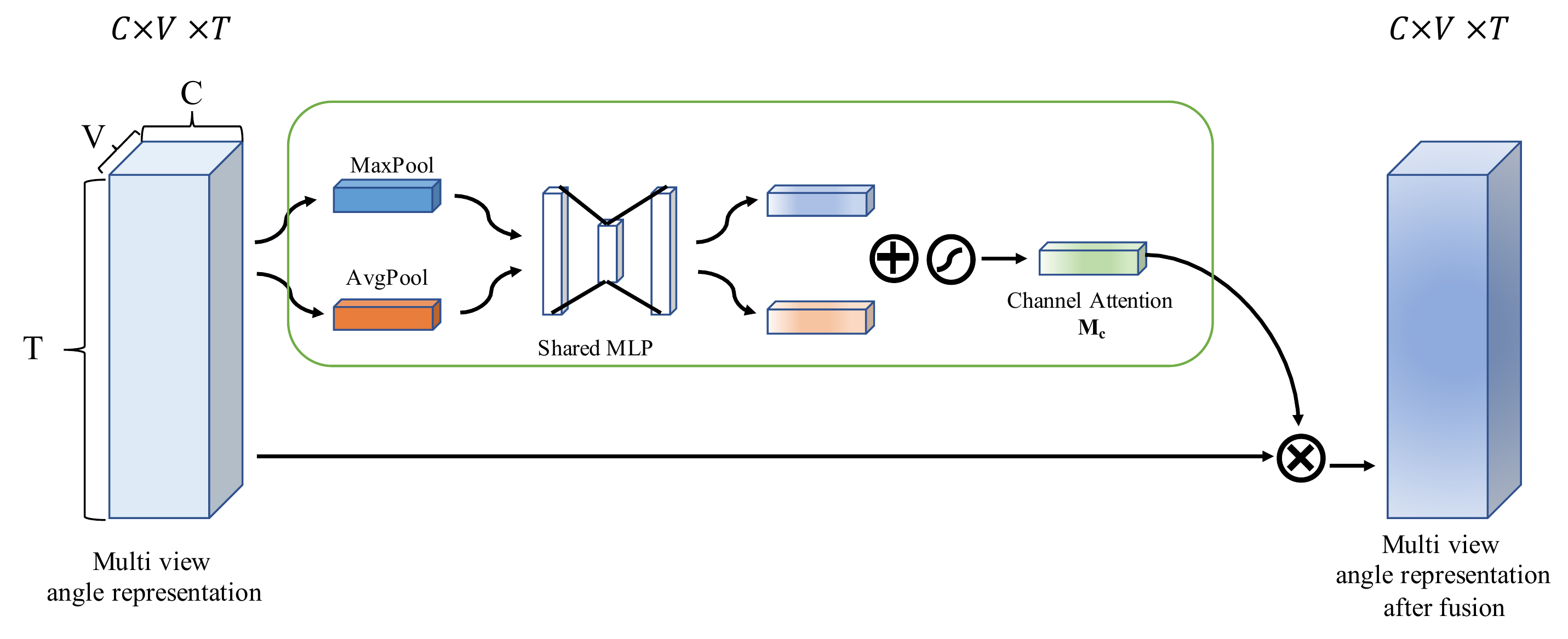}
    % \caption{Aggregation of the multi view angle representation. We operate average pooling and max pooling in the joints and time dimensions respectively. Then perform squeeze and excitation on the two tensors generated before. Then sum the two tensors and pass the activation function to get the channel attention factor . Finally, multiply the factors with the input to aggregate the views.}
    \caption{Aggregation of the multi view angle representation.}   
    \Description{We illustrate the aggregation of the multi view angle representation. We operate average pooling and max pooling in the joints and time dimensions respectively. Then perform squeeze and excitation on the two tensors generated before. Then sum the two tensors and pass the activation function to get the channel attention factor . Finally, multiply the factors with the input to aggregate the views.}
    \label{fig:fusion}
\end{figure}

After the angle representation for each view is obtained with the learned corresponding pair of anchors, the next step is to aggregate those multiple view information into the following GCN module for feature extraction. We employ a channel-wise attention mechanism that is inspired by \cite{woo2018cbam} in this stage as shown in Figure \ref{fig:fusion}. We operate average pooling and max-pooling in the joints and time dimensions respectively. Then perform squeeze and excitation on the two generated tensors. Then summation result of two tensors is passed the activation function to obtain the channel attention factor. Finally,  origin multi-view angle representation is multiplied by the channel factor to conduct the fusion of multi-view information. Inspired by existing multi-feature fusion works, we also explore a wide range of different fusion strategies, which would be shown in the ablation study. 

\section{Experimental Results and Discussions}
\begin{table*}[htbp]
  \centering
%   \begin{threeparttable}
  \caption{Comparison with State-of-the-Art models}
    \begin{tabular}{lccccccc}
    \toprule
    \multicolumn{1}{c}{\multirow{2}[3]{*}{Methods}} & \multicolumn{1}{c|}{\multirow{2}[3]{*}{Publisher}} & \multicolumn{2}{c|}{NTU60} & \multicolumn{2}{c|}{NTU120} & \multicolumn{2}{c}{Kinetics Skeleton 400} \\
\cmidrule{3-8}          & \multicolumn{1}{c|}{} & \multicolumn{1}{c|}{X-Sub (\%)} & \multicolumn{1}{c|}{X-View (\%)} & \multicolumn{1}{c|}{X-Sub (\%)} & \multicolumn{1}{c|}{X-Set (\%)} & \multicolumn{1}{c|}{Top-1 (\%)} & Top-5 (\%) \\\midrule
    ST-GCN\cite{yan2018spatial} & AAAI18 & 81.5  & 88.3  & -     & -     & 30.7  & 52.8 \\
    2s-AGCN\cite{shi2019two} & CVPR19 & 88.5  & 95.1  & 82.9  & 84.9  & 36.1  & 58.7 \\
    DGNN\cite{shi2019skeleton}  & CVPR19 & 89.9  & 96.1  & -     & -     & 36.9  & 59.6 \\
    DSTA-Net\cite{shi2020decoupled} & ACCV20 & 91.5  & 96.4  & 86.6  & 89    & -     & - \\
    DDGCN\cite{korban2020ddgcn} & ECCV20 & 91.1  & 97.1  & -  & -    & 38.1     & 60.8 \\
    4s Shift-GCN\cite{cheng2020skeleton} & CVPR20 & 90.7  & 96.5  & 85.9  & 87.6  & -     & - \\
    MST-GCN\cite{chen2021multi} & AAAI21 & 91.5  & 96.6  & 87.5  & 88.8  & 38.1  & 60.8 \\
    MG-GCN\cite{chen2021learning} & ACMMM21  & 92    & 96.6  & 88.2  & 89.3  & 38.4  & 61.3 \\
    EfficientGCN\cite{song2022constructing} & TPAMI22 & 91.7  & 95.7  & 88.3  & 89.1  & -     & - \\
    AngNet\cite{qin2021fusing} & arxiv22 & 91.6  & 96.3  & 88.2  & 89.2  & -     & - \\
    VA-CNN\cite{zhang2019view} & TPAMI19 &88.7 &94.3 & - & - & - & - \\
    MS-G3D\cite{liu2020disentangling} & CVPR20 & 91.5  & 96.2  & 86.9  & 88.4  & 38    & 60.9 \\
    CTR-GCN\citep{chen2021channel}  & ICCV2021 & 92.7 & 96.8 & 88.9  & 90.6 & -     & - \\
    \midrule
    \multicolumn{8}{c}{Our Methods} \\
    \midrule
    SAP+VA-CNN &- & 89.1\textbf{(+0.4)} & 95.1\textbf{(+0.8)} & - &- &- &-\\
    SAP+CTR-GCN & -     & 93.0\textbf{(+0.3)}  & 96.8  & 89.5\textbf{(+0.6)}  & 91.1\textbf{(+0.5)}  & -     & - \\
    SAP+MS-G3D & -     & 92.7\textbf{(+1.2)}  & 97.0\textbf{(+0.8)}  & 88.8\textbf{(+1.9)}  & 90.4\textbf{(+2.0)}  & 38.8\textbf{(+0.8)} & 61.7\textbf{(+0.8)} \\
    \bottomrule
    \end{tabular}%
  \label{tab:sota}%
\end{table*}%
\begin{table}[htbp]
  \centering
  \caption{Robustness Study with noisy data}
  \begin{tabular}{lcc}
        \toprule
        \multicolumn{1}{c}{Methods}       & MS-G3D & MS-G3D+SAP \\
        \midrule
        Original Data        & 91.5   & 92.7       \\
        \hline \multicolumn{3}{c}{Random rotation} \\\hline
        \lbrack -0.1 rad, 0.1 rad \rbrack    & 91.2   &  92.4      \\
        \lbrack -0.2 rad, 0.2 rad \rbrack    & 90.8   &  92.3     \\
        \lbrack -0.3 rad, 0.3 rad \rbrack    & 89.9   &  92.0     \\
        \hline \multicolumn{3}{c}{Random remove joint} \\\hline
        remove 1 joint in 10\% frames  &   91.2 & 92.4      \\
        remove 10 joints in 10\% frames  & 90.1   & 91.3      \\
        remove 15 joints in 10\% frames  & 89.5   & 90.6      \\
        \hline \multicolumn{3}{c}{Random disturb} \\\hline
        disturb 1 joint in 1\% frames    & 91.2   & 92.5       \\
        disturb 10 joints in 1\% frames  & 89.7   & 92.1       \\
        disturb 25 joints in 1\% frames & 88.1   & 91.7       \\
        \bottomrule
\end{tabular}

  \label{tab:robust}%
\end{table}%
\begin{table*}[htbp]
    
    \begin{subtable}[t]{0.33\textwidth}
    \centering
    \caption{Number of View Anchors}
  {
   
  \begin{tabular}{lcc}
    \toprule
    \multicolumn{1}{c|}{\multirow{2}[4]{*}{Pairs of View Anchors}} &
    \multicolumn{2}{c}{NTU60} \\
\cmidrule{2-3}  \multicolumn{1}{c|}{} & X-Sub & Acc $\uparrow$ \\
    \midrule
    fixed 7      & 77.1  & - \\
    1     & 75.84 & - \\
    3    & 82.38 & 5.28 \\
    \textbf{5}       & \textbf{84.7}  & \textbf{7.6} \\
    7      & 84.32 & 7.22 \\
    10      & 84.27 & 7.17 \\
    15      & 84.81 & 7.71 \\
    \bottomrule
    \end{tabular}
  \label{tab:ablation-num-head}
  }
        
    \end{subtable}
    \begin{subtable}[t]{0.33\textwidth}
        \centering
        \caption{Effectiveness of the different fusion strategy}
    \begin{tabular}{cc}
    \toprule
    \multicolumn{1}{c|}{\multirow{2}{*}{Fusion methods}} & NTU60 \\
\cmidrule{2-2}    
\multicolumn{1}{c|}{} 
& X-Sub (\%) \\ 
\midrule
    SUM   & 76.9 \\
    MAX   & 75.9 \\
    CONCATENATE   & 84.4 \\
    ATTENTION & \textbf{84.7}\\
    \bottomrule
    \end{tabular}%
  \label{tab:ablation-fusion}%
    \end{subtable}
    \begin{subtable}[t]{0.33\textwidth}
      \centering
      \caption{Anchor Location Constraint}
  {
    \begin{tabular}{lcc}
    \toprule
    \multicolumn{1}{c|}{\multirow{2}[4]{*}{Anchor Location}} &
    
    \multicolumn{2}{c}{NTU60} \\
\cmidrule{2-3}
\multicolumn{1}{c|}{} & X-Sub & Acc $\uparrow$ \\
    \midrule
    fixed 7 pairs of Joints     & 77.1  & - \\

    Anchors on Joints      & 78.5  & 1.4 \\
    Anchors within Body      & 78.7  & 1.6 \\
    
    Anchors around Body      & \textbf{84.7}  & \textbf{7.6} \\
    \bottomrule
    \end{tabular}%
  \label{tab:ablation-anchor-location}%
  }
    \end{subtable}
    \caption{Ablation study of SAP module: (a) Effect of Number of Views, (b) Effectiveness of the different fusion strategy and (c) Effectiveness of Anchor Location Constrain}
    \label{tab:array}
\end{table*}
\label{Exp}
In this section, we carried out extensive comparisons and ablation experiments to demonstrate the effectiveness, generality and robustness of our proposed idea. 
% The details of our network configuration are given out in the section. 
Comprehensive quality and quantity results are reported to show the superiority of the multi-view strategy against the state-of-the-art models. Discussions of these experiments point out the interesting findings of our work. 

\subsection{Implementation Details.}
\noindent\textbf{Network Settings.}  The entire model has been trained on 2 NVIDIA RTX 3090 GPUs with PyTorch. Stochastic gradient descent (SGD) is applied with a momentum of 0.9 and a learning rate of 0.05 with a step of 10 times decay at the 30th and 40th epochs. For simplicity, a modified Resnet-50 is used as backbone model to determine the hyper-parameter in the ablation study. After that, our SAP head is installed with the baseline models for the fine-tuning process to further enhance the model performance.

\noindent\textbf{Dataset and Metric.} 
The performance of proposed method is evaluated on three large-scale public skeleton-based datasets: NTU-RGBD 60 ~\cite{shahroudy2016ntu}, NTU-RGBD 120 \cite{liu2020disentangling} and Kinetics-Skeleton \cite{yan2018spatial}. 
\begin{itemize}
    \item \textbf{NTU RGBD 60} dataset contains 60 different human action classes. It consists of 56,880 action samples in total which are performed by 40 distinct subjects. The 3D skeleton data is collected by Microsoft Kinect v2 from three cameras simultaneously with different horizontal angles: $-45, 0, 45$. The human pose in each frame is represented by $25$ joints.  
    \item \textbf{NTU RGBD 120} dataset is an extended version of the NTU-RGBD 60 dataset by adding another 60 classes and another 57,600 video/skeleton samples. It consists of 114,480 action samples divided into 120 action classes.
    \item  \textbf{Kinetics-Skeleton} is an activity recognition dataset for skeleton-based action recognition, which consists of 300,000 clips in 400 classes. The training data is set to 240,000 skeleton clips, and the test data consists of 20,000 clips. 
\end{itemize}
We follow two official evaluation protocols for performance evaluation for NTU RGBD datasets: Cross-Subject (X-Sub) and Cross-View (X-View). For kinetics-Skeleton dataset, we use both the top-1 and top-5 accuracy as other methods do. 

\noindent\textbf{Baseline Models.} 
We have employed 
% VA-CNN~\cite{zhang2019view}, and x
MS-G3D~\cite{liu2020disentangling} as our baseline and the backbone for the action recognition task, where MS-G3D is a typical work that introduces multi-scale graph topologies to GCNs to enable multi-range joint relationship modelling \cite{liu2020disentangling}. The primary focus of this paper is to validate the proposed SAP's ability to provide complementary information to current popular models. Current SOTA CTR-GCN \cite{song2022constructing} hasn't been chosen as a baseline model since its performance gain is largely benefited from its data augmentation strategy while other methods do not take this operation. 

\subsection{Compared with the state-of-the-art methods}

\noindent\textbf{Effectiveness}
We compare our method against existing SOTA models on both X-Sub and X-View benchmarks. The results are presented in Table \ref{tab:sota}. For NTU RGBD 60, the accuracy of our proposed SAP with MS-G3D on the X-Sub benchmark is $92.7\%$ and the X-view benchmark is $97.0\%$, which outperforms most other trending action recognition models. Compared with the baseline model MS-G3D~\cite{liu2020disentangling}, our work brings a marginal improvement on both X-Set and X-Sub. Similarly, an evident improvement of proposed method has occurred on the Kinetics Skeleton dataset. 

Compare with current SOTA work CTR-GCN\cite{chen2021channel}, we employ its data augmentation prepossessing operation and additional velocity branch,  the performance still has a gain.

These results manifest that the proposed SAP module is able to conduct a competitive performance compared to the existing best models. We consider that our method is able to provide complementary information via exploiting the multi-scale relative motion pattern via encoding different view information. Moreover, the attention mechanism is able to encourage the model to discover the informative joints with a complementary view.

\noindent\textbf{Transferability}
We evaluate if our method is effective for diverse types of action recognition models. 
% Working together with the best performance model CTR-GCN~\cite{chen2021channel}, our SAP module still lifts the accuracy to be slightly higher and achieves the Rank-1 performance on the existing benchmark. 
To validate that our SAP module is also effective for CNN backbone models, we conduct our multi-view strategy on the VA-CNN (Resnet-50 version)~\cite{zhang2019view} models. A significant enhancement can be introduced with our SAP module. As reported in the Table\ref{tab:sota}, the SAP module boosts the performance about 0.4 on X-Sub. These results demonstrate that our method is unique to both GCN-based and CNN based backbones. 

\noindent\textbf{Robustness}
The robustness is evaluated with 3 kinds of synchronized noise data that contains missing and corrupted joints: (1) Rotated Data. We rotate the entire action sequence with a certain noisy angle to change the initial view direction of the training data. (2) Noisy Data. We introduce Gaussian noise disturb with a mean of 0 and a variance of 1 to the randomly sampled $1\%$ frames. This case is similar to the real-life scenarios in which the motion data contains a lot of noise. (3) Missing Data. We remove some value of joints from randomly sampled $10\%$ frames to analyse the model performance with missing bones or corrupted data. MS-G3D~\cite{liu2020disentangling} is used as the baseline model. The results are shown in Table \ref{tab:robust}, it is clearly validated that our model can substantially reduce the effects of noise problems.

\subsection{Ablation Study}
All experiments in this section are conducted on the NTU-60 dataset for the purpose of analysing the different components and designs of our method. The overall test accuracy of experiments is compared following the linear evaluation protocol of Cross-Subject (X-sub) while Resnet-50 is chosen as the baseline model. The training details are available in the public repository. 

\noindent\textbf{Effect of Number of Views.} 
We evaluate the relationship between the number of views and the final performance gains. As shown in Table \ref{tab:ablation-num-head}, we variate the number $M$ of multi-head designs in our SAP module to generate $M$ different pairs of view anchors. The selection range is from $1$ to $15$. The first row indicates the baseline which uses manually specified 7 pairs of joints as anchors that are used in \cite{qin2021fusing}. The result has shown that dynamically learned anchors from the proposed SAP significantly outperform the fixed ones in most cases. The gains become smaller after 5. Considering the trade-off between effectiveness and efficiency, we finally select a 5-head structure for further performance evaluation.

\noindent\textbf{Effectiveness of the different fusion strategy}
We analyse four typical fusion strategies in the pipeline to figure out the best way to aggregate the view features, which is shown in Table~\ref{tab:ablation-fusion}. 
From these results, we can conclude that the different fusion strategies actually affect the final performance significantly with a large range of 8.8 on X-Sub. Attention fusion achieves the highest accuracy 84.7 on X-Sub, which is finally used in this work.

\noindent\textbf{Effectiveness of Anchor Location Constraint.} Hyper-parameters $\alpha$ and $g(x)$ are used to control the range of anchor location. For instance, the anchors would be limited on joints with a large $\alpha$, e.g. $20$ in our experiment. An example of anchors located around the body is shown in Figure \ref{fig:AnchorPoints}. We noticed that the learning anchors with less limitation could achieve better performance since it can provide bigger receptive fields .

\section{Conclusion}
\label{con}
In this paper, we address the skeleton-based action recognition task from view enhancement strategy by proposing a model-agnostic Skeleton Angle Proposal (SAP) module together with a new angle representation. For the first time, the view information for skeleton-based action data has been in-depth investigated and analysed. Our extensive experiments reveal that this idea could boost performance by a significant margin. It is worth noting that, our SAP module is generic and robust to seamlessly work with any existing model. Apart from these validations, systematical ablation studies are carried out to figure out the best choices for the number of views and positions of anchors. Comprehensive experiments also contribute fruitful insights into the model design aspect. In contrast to the existing approaches, this paper is a first step to understanding actions from a view aspect and the promising results will encourage future research to open a new trend in the action recognition field. We expect our new techniques could have great potential to facilitate action recognition research and benefit industrial-level applications. 

\begin{acks}
% \noindent\textbf{Acknowledgements.}
This research has been supported by National Key Research \& Development Project of China (2021ZD0110700), National Natural Science Foundation of China (U19B2042), Zhejiang Provincial Natural Science Foundation of China (LGG22F030011), Ningbo Natural Science Foundation (2021J167,2021J190,2022Z072), Yongjiang Talent Introduction Programme (2021A-156-G), the Cloud based Virtual Production Project funded by the Arts and Humanities Research Council (UK-AHRC Ref: AH/W009323/1) and the Neuravatar Project funded by Higher Education Innovation Fund (UK-HEIF).
\end{acks}

%% References
\bibliographystyle{ACM-Reference-Format}
\bibliography{sample-base}

\end{document}